\documentclass[10pt,twocolumn,letterpaper]{article}

\usepackage{cvpr}              

\usepackage{graphicx}
\usepackage{amsmath}
\usepackage{amssymb}
\usepackage{booktabs}
\usepackage{multicol}
\usepackage{multirow}
\usepackage{bbding}
\usepackage[pagebackref,breaklinks,colorlinks]{hyperref}

\usepackage[capitalize]{cleveref}
\crefname{section}{Sec.}{Secs.}
\Crefname{section}{Section}{Sections}
\Crefname{table}{Table}{Tables}
\crefname{table}{Tab.}{Tabs.}

\def\cvprPaperID{*****} 
\def\confName{CVPR}
\def\confYear{2023}

\def\ie{\emph{i.e.,~}}
\def\eg{\emph{e.g.,~}}

\newcommand{\myPara}[1]{\vspace{.05in}\noindent\textbf{#1.}\quad}

\newlength\savedwidth
\newcommand{\whline}[1]{\noalign{\global\savedwidth\arrayrulewidth \global\arrayrulewidth #1}%
                   \hline \noalign{\global\arrayrulewidth\savedwidth}}

\begin{document}

\title{Rethinking the Open-Loop Evaluation of End-to-End Autonomous Driving in nuScenes}

\author{
Jiang-Tian Zhai\thanks{Equal contribution.} \quad
Ze Feng\footnotemark[1] \quad
Jinhao Du\footnotemark[1] \quad
Yongqiang Mao\footnotemark[1] \quad
Jiang-Jiang Liu\thanks{Project lead.}\\
Zichang Tan \quad
Yifu Zhang \quad
Xiaoqing Ye \quad
Jingdong Wang\footnotemark[2] \\ 
[2mm]
Baidu Inc. \\
\tt\small{\{jtzhai30, j04.liu\}@gmail.com,
wangjingdong@outlook.com}
}

\maketitle

\begin{abstract}
Modern autonomous driving systems are typically divided into three main tasks: perception, prediction, and planning. 
The planning task involves predicting the trajectory of the ego vehicle 
based on inputs from both internal intention and 
the external environment, and manipulating the vehicle accordingly. 
Most existing works evaluate their performance on the nuScenes dataset using the L2 error and collision rate between the predicted trajectories and the ground truth. 
In this paper, we reevaluate these existing evaluation metrics and explore whether they accurately measure the superiority of different methods. 
Specifically, we design an MLP-based method that takes raw sensor data (e.g., past trajectory, velocity, etc.) as input and directly outputs the future trajectory of the ego vehicle, 
without using any perception or prediction information such as camera images or LiDAR. 
Our simple method achieves similar end-to-end planning performance on the nuScenes dataset with other perception-based methods, 
reducing the average L2 error by about 20\%. 
Meanwhile, the perception-based methods have an advantage in terms of collision rate.
We further conduct in-depth analysis and provide new insights into the factors that are critical for the success of the planning task 
on nuScenes dataset.
Our observation also indicates that we need to rethink the current open-loop evaluation scheme
of end-to-end autonomous driving in nuScenes.
Codes are available at \url{https://github.com/E2E-AD/AD-MLP}.

\end{abstract}

\section{Introduction}
\label{sec:intro}
Many existing autonomous driving models~\cite{liang2020pnpnet,luo2018fast,sadat2020perceive} involve a multi-stage pipeline of independent tasks, such as perception\cite{Xiong2023CAPE,li2022bevformer}, prediction\cite{fiery2021,densetnt} and planning\cite{Chitta2022PAMI,chen2021}. While this design simplifies the difficulty of collaboration across teams, it leads to information loss and error accumulation in the overall system due to the independence of optimization targets and model training. To better predict control signals and enhance user safety, an end-to-end approach that benefits from spatial-temporal feature learning from the ego vehicle and surrounding environment is desired.

\begin{figure*}[ht]
    \centering
    \small
    \includegraphics[width=1.0\linewidth]{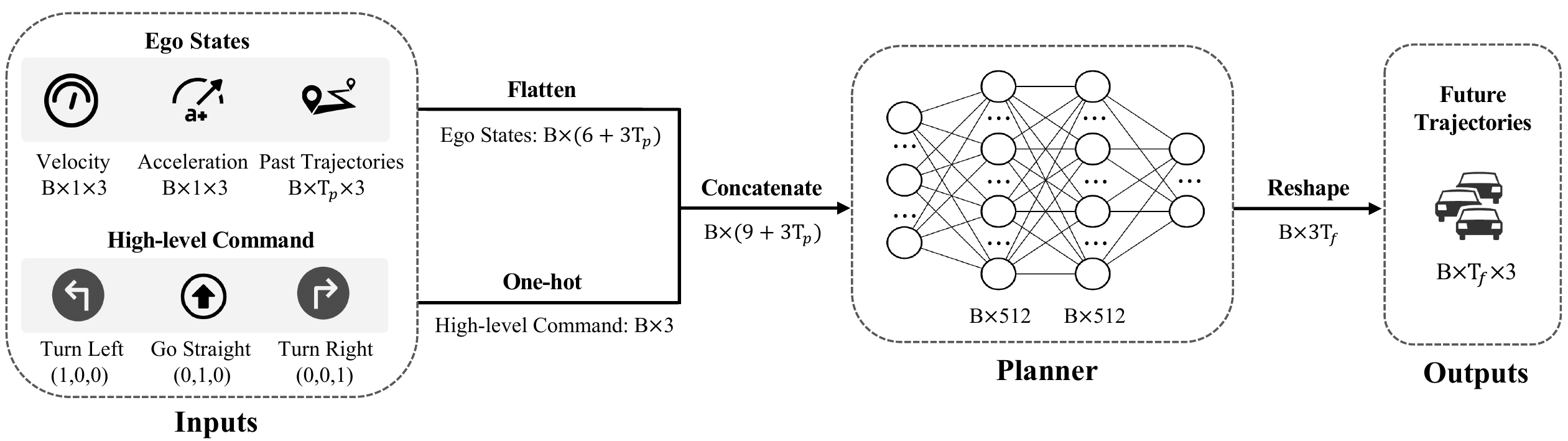}
    \caption{
    Overall pipeline. 
    \textbf{Inputs:}
    1) ego vehicle's states: 
    motion trajectories ($x,y,\theta$) of the past $T_p$ frames (in our experiments $T_p=4$), 
    instantaneous velocity ($v_x,v_y,\omega$)
    and acceleration ($a_x,a_y,\beta$). $\theta$, $\omega$, and $\beta$ indicate heading angle, angular velocity, and angular acceleration, respectively;  
    \enspace2) high-level command (one-hot encoded). $B$ indicates batch size.
    \textbf{Outputs:}
    motion trajectory of the ego vehicle in the 
    future $T_f$ frames (in our experiments $T_f=6$).
    }
    \label{fig:Method}
\end{figure*}

Planning for autonomous driving is the ultimate goal of the entire system. To achieve these, some methods are accomplished through perception tasks, such as 3D object detection and semantic segmentation to obtain temporal and spatial information about the surrounding environment. 
A natural idea is that if a model can perform well in these perception tasks, it can make accurate, safe, and comfortable trajectory planning based on this information. ST-P3~\cite{hu2022st} proposes an interpretable vision-based end-to-end system that unifies feature learning for perception, prediction, and planning. 
UniAD~\cite{hu2022goal} adopts a systematic model design for the planning task by connecting all intermediate task nodes based on the query-based design where the relationship between multiple tasks can be modeled and encoded. VAD~\cite{jiang2023vad} models the scene in a fully vectorized way, getting rid of computationally intensive grid feature representation and being more efficient in computation. 
This vectorized feature representation helps autonomous driving vehicles focus on the crucial map and agent elements, and plan a reasonable future trajectory.
The above-mentioned methods 
achieve promising performance on the trajectory planning task of the ego vehicle on the nuScenes~\cite{caesar2020nuscenes} dataset, which is the most commonly used one
in the area.

However, in this paper, in order to explore whether the existing evaluation metrics can accurately measure the superiority of different methods, we only use the physical state of the ego vehicle during driving as input (\ie a subset of the information used by existing methods) to conduct experiments, instead of using the perception and prediction information provided by the camera or LiDAR.
In other words, there is  
no encoder for visual or point cloud feature extraction 
in the model.
We directly unfold all the physical information of the ego vehicle into one-dimensional vectors and feed them into
a multi-layer perceptron (MLP)
after concatenation.

During training, we use the ground-truth trajectory as supervision and have the model directly predict the ego vehicle's future trajectory points for a certain period of time. 
Following previous methods~\cite{hu2022st,hu2022goal}, 
we validate our approach on the nuScenes dataset using the metrics of L2 error and collision rate.

Although the design of our model is simple and no perception information is utilized, 
it achieves similar trajectory planning performance on the nuScenes dataset.
We attribute this to the inadequacy of current evaluation metrics for planning tasks on the nuScenes dataset to accurately compare the performance of different methods.
In fact, the motion of the ego vehicle in the future can be reflected to a certain extent by using the past trajectory, velocity, acceleration information, and temporal continuity. 
%


\section{Method}
The overall pipeline of our method is illustrated in Figure~\ref{fig:Method}. 
As our model requires no perception information, 
there is no component for 
visual or point-cloud feature extraction. 
The inputs only consist of two parts: the ego states
and the high-level command representing its future short-term motion trend. 
Our model is trained in an end-to-end manner.

\subsection{Model Inputs}
\myPara{Ego States} 
Following~\cite{jiang2023vad}, 
we collect the ego vehicle's motion trajectories of the past $T_p=4$ frames, instantaneous velocity and acceleration as input.
The trajectory of each frame consists of 
three values: ($x,y,\theta$), representing 
the position and heading angle of ego vehicle, respectively. 
The instantaneous velocity ($v_x,v_y,\omega$) 
and acceleration ($a_x,a_y,\beta$) represent
the current $x$-directional, $y$-directional, and angular velocity and acceleration, respectively.
We obtain this information from the nuScenes CAN bus expansion. 
The above-mentioned ego states
are then flattened and concatenated into a 
one-dimensional vector. 

\myPara{High-Level Command}
Since our model does not use HD maps as input, a high-level command is required for navigation. 
Following the common practice~\cite{hu2022st,hu2022goal}, three types of commands are defined: 
turn left, go straight, and turn right. 
Specifically, when the ego vehicle displaces $>2m$ to the left or right direction in the future 3s, the corresponding command is set to turn left or right. 
Otherwise, it corresponds to going straight. 
We represent the command using a $1 \times 3$ one-hot encoding.  

The ego states and high-level command are 
concatenated together as the network's inputs, so the final dimension of the input vector is 21.

\myPara{Network Structure}
The structure of our model is a simple
MLP 
and can be summarized as $Linear_{512}^{21}-ReLU-Linear_{512}^{512}-ReLU-Linear_{18}^{512}$.
For each $Linear_{C_{in}}^{C_{out}}$, 
$C_{in}$ and $C_{out}$ represent the numbers of 
input and output channels, respectively.
The final outputs represent the ego vehicle's trajectories (x,y coordinates) and heading angles for the future $T_f=6$ frames.

\begin{table*}
    \centering
    \small
    \renewcommand{\arraystretch}{1.2}
    \setlength\tabcolsep{1.52mm}
    \begin{tabular}{c|c|c|ccc|ccc|c|ccc|c}
    \whline{1pt}
    \multirow{2}*{Method} & Perception &  High-level & \multicolumn{3}{c|}{Ego State}  & \multicolumn{4}{c|} {L2 (m) $\downarrow$} & \multicolumn{4}{c}{Collision (\%) $\downarrow$} \\ 
    \cline{4-14}
      & Information & Command & Velocity & Acceleration & Trajectory & 1s & 2s & 3s & \textbf{Avg.} & 1s & 2s & 3s & \textbf{Avg.} \\
    \whline{1pt}
    NMP~\cite{zeng2019end}     & \Checkmark &\XSolidBrush & - & - & - &  -   &  -   & 2.31 &  -   &  -   &  -   & 1.92 &  -   \\
    SA-NMP~\cite{zeng2019end}  & \Checkmark &\XSolidBrush & - & - & - &  -   &  -   & 2.05 &  -   &  -   &  -   & 1.59 &  -   \\
    FF~\cite{hu2021safe}       & \Checkmark &\XSolidBrush & - & - & - & 0.55 & 1.20 & 2.54 & 1.43 & 0.06 & 0.17 & 1.07 & 0.43 \\
    EO~\cite{khurana2022differentiable}  & \Checkmark &\XSolidBrush & - & - & - & 0.67 & 1.36 & 2.78 & 1.60 & 0.04 & 0.09 & 0.88 & 0.33 \\
    \whline{0.5pt}
    ST-P3~\cite{hu2022st}        & \Checkmark &\Checkmark & - & - & - & 1.33 & 2.11 & 2.90 & 2.11 & 0.23 & 0.62 & 1.27 & 0.71 \\
    UniAD~\cite{hu2022goal}      & \Checkmark &\Checkmark& - & - & - & 0.48 & 0.96 & 1.65 & 1.03 & 0.05 & 0.17 & 0.71 & 0.31 \\
    VAD-Tiny~\cite{jiang2023vad} & \Checkmark &\Checkmark & \Checkmark & \Checkmark & \Checkmark & 0.20 & 0.38 & 0.65 & 0.41 & 0.10 & 0.12 & 0.27 & 0.16 \\
    VAD-Base~\cite{jiang2023vad} & \Checkmark &\Checkmark & \Checkmark & \Checkmark & \Checkmark & 0.17 & 0.34 & 0.60 & 0.37 & 0.07 & 0.10 & 0.24 & 0.14 \\
    \whline{0.5pt}
    \multirow{4}*{Ours}  & \XSolidBrush & \XSolidBrush & -  & - & \Checkmark & 0.53 & 0.91 & 1.48 & 0.97 & 0.17 & 0.46 & 0.83 & 0.49 \\
    ~  & \XSolidBrush & \XSolidBrush & -  & \Checkmark & \Checkmark & 0.33 & 0.48 & 0.66 & 0.49 & 0.21 & 0.29 & 0.40 & 0.30 \\
    ~  & \XSolidBrush & \XSolidBrush & \Checkmark  & \Checkmark & \Checkmark & 0.24 & 0.32 & 0.49 & 0.35 &  0.18 & 0.22 & 0.28 & 0.23 \\
     & \XSolidBrush & \Checkmark & \Checkmark  & \Checkmark & \Checkmark & 0.20 & 0.26 & 0.41 & 0.29 & 0.17 & 0.18 & 0.24 & 0.19 \\
    \whline{1pt}
    \end{tabular}
    \caption{Comparison with existing perception-based methods. 
    Our method achieves slightly lower L2 error, while the collision rate is higher than some other methods.
    Results in the table except for our method are collected from VAD~\cite{jiang2023vad}.
    }
    \label{performance}
\end{table*}
\subsection{Loss Function}
The loss function $\mathcal{L}$ is implemented with a L1 loss between the predicted planning trajectories $\hat{V}_{ego}$ and the ground-truth ego trajectories $V_{ego}$, which can be formulated as follows: 
\begin{equation}
  \begin{aligned}
    \label{eqn:cel}
    \mathcal{L} &=   \frac{1}{N_f} \sum_{i=1}^{N_f} ||\hat{V}_{ego}-V_{ego}||_1
  \end{aligned}
\end{equation}
where $N_f$ denotes the number of future frames, \eg{6}. 
Since the resolution of occupancy map is 0.5m for one grid, the predicted trajectory point is mapped to the same grid if it falls into the same 0.5m segment, \eg{$[1.5m,2m)$}, with the ground truth on both x and y axes. 
We re-weight the loss by 0.5 on samples where the predicted trajectory points and the ground truth coincide at the same grid for hard sample mining to reduce the collision rate. 

\section{Experiments}
\subsection{Dataset \& Evaluation Metrics}
\myPara{Dataset} 
Following the common practice~\cite{hu2022st,hu2022goal,jiang2023vad}
in the planning task,
we use the nuScenes~\cite{caesar2020nuscenes}
dataset in our experiments for both training and testing.
The dataset includes 1K scenes and approximately 40K key-frames 
mainly collected in Boston and Singapore 
using vehicles equipped with both LiDAR and surrounding cameras. 
The data collected for each frame includes multi-view camera images, LiDAR, velocity, acceleration, etc.

\myPara{Metrics}
We use the implementation\footnote{\url{https://github.com/OpenPerceptionX/ST-P3/blob/main/stp3/metrics.py}} provided by ST-P3\cite{hu2022st} to evaluate the output trajectories for time horizons of 1s, 2s, and 3s. 
To evaluate the quality of the predicted ego trajectories, 
two commonly used metrics~\cite{hu2022st,hu2022goal,jiang2023vad} are calculated: 
L2 error (in meters) and collision rate (in percentage).
The average L2 errors are calculated between the predicted and ground-truth trajectories for corresponding waypoints within the next 1s, 2s, and 3s time horizons, respectively. 
To determine how often the ego vehicle 
collides with other objects, 
the collision rate is computed by 
placing a box representing the ego vehicle at each waypoint on the predicted trajectory 
and then detecting if any collision with other oriented bounding boxes that represent vehicles and pedestrians in the scene occurs.

\subsection{Implementation Details} 
Our model is implemented in both the PaddlePaddle and PyTorch framework.
The AdamW~\cite{loshchilov2017decoupled} optimizer is used with an initial learning rate of 4e-6 and weight decay of 1e-2. 
The cosine annealing~\cite{loshchilov2016sgdr} 
learning rate schedule is utilized.
Our model is trained for 6 epochs 
with a batch size of 4 
on 1 NVIDIA Tesla V100 GPUs.

\newcommand{\addts}[1]{\includegraphics[height=0.225\linewidth]{fig/#1.pdf}}
\begin{figure*}[t]
    \centering
    \small
    \setlength\tabcolsep{0.2mm}
    \renewcommand\arraystretch{0.6}
    \begin{tabular}{ccc}
        \addts{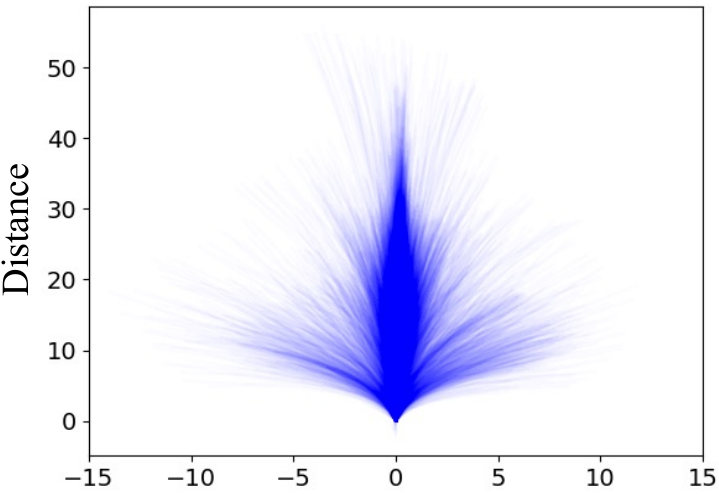} & 
        \addts{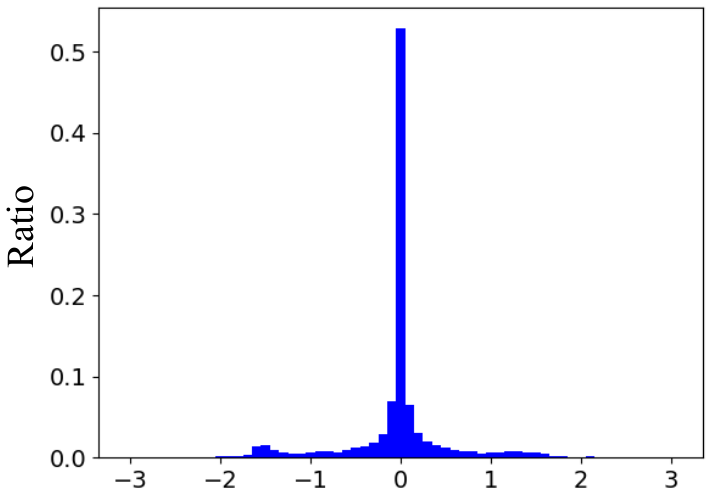} & 
        \addts{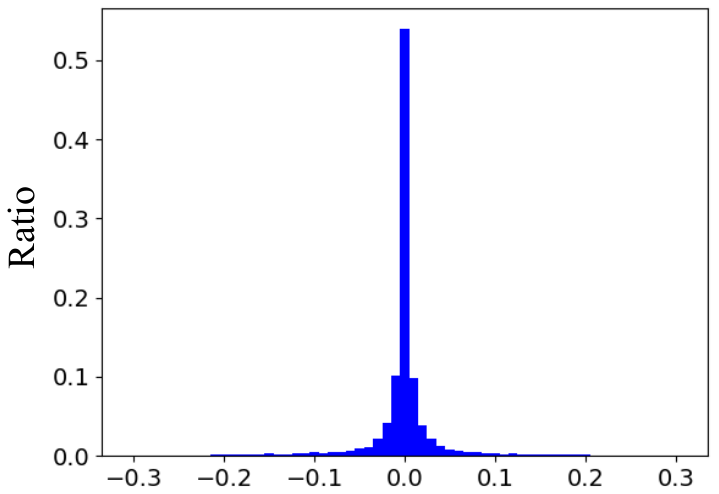}
        \vspace{4pt} \\
        \quad\quad (a) Trajectory Points & \quad\quad (b) Heading Angle & \quad\quad (c) Curvature Angle\\
    \end{tabular}
    \caption{
    Distribution analysis of nuScenes training set.
    The trajectory points are concentrated in the middle forward area, and the heading angle and curvature angle are concentrated around radian 0. We can conclude that most cases of the ego vehicle are in straight and small angles forward, and there are few cases of large angle turns.
    }
    \label{fig:Distribution}
\end{figure*}

\begin{figure*}[t]
    \centering
    \small
    \includegraphics[scale=0.245]{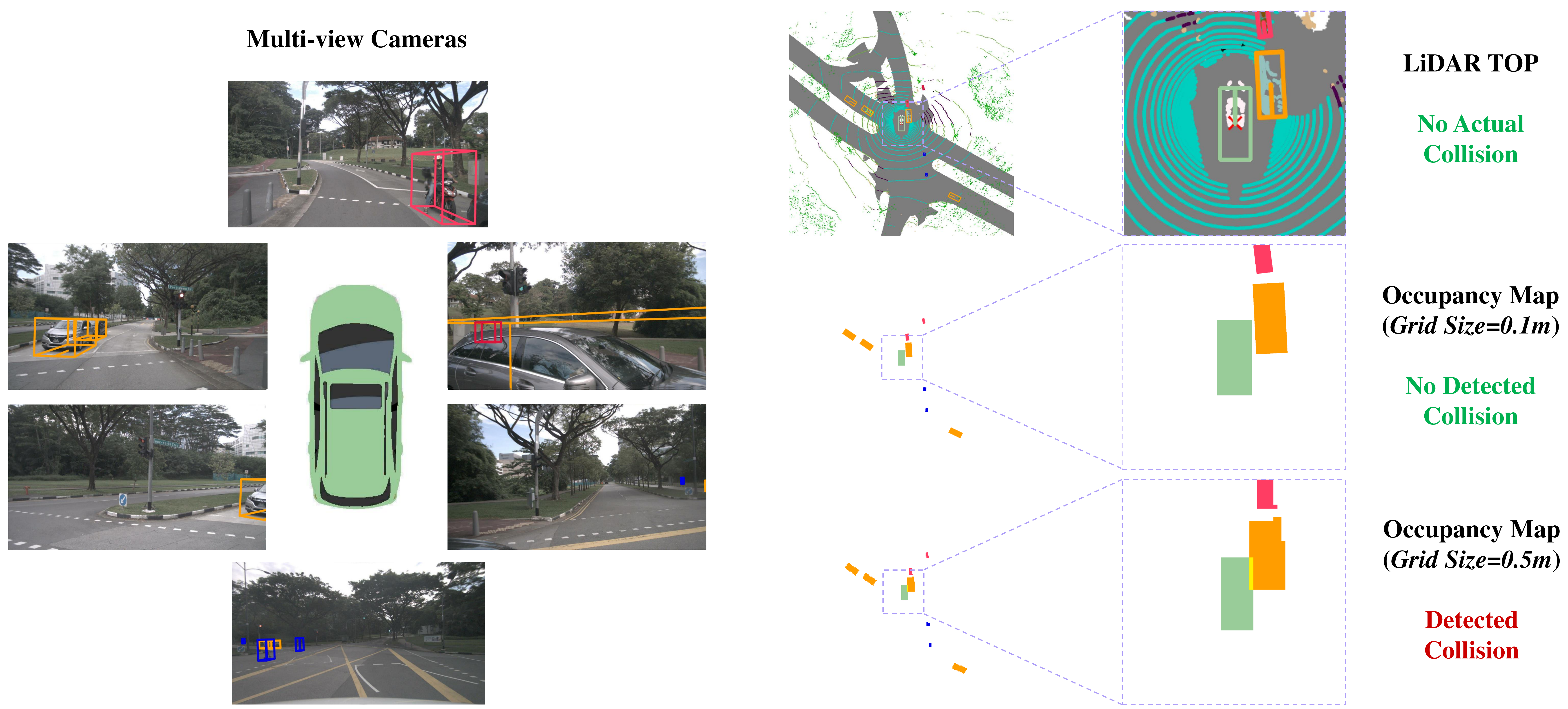}
    \caption{
    A typical ground-truth trajectory collision case caused by different grid sizes of the occupancy maps.
    It can be observed that the gird size for occupancy map generating has a great impact on the collision test, 
    which is commonly used in the evaluation of collision in 
    existing methods.
    For instance, in the case of $grid~size=0.1m$, the ground-truth trajectory is correctly recognized as a no-collision case while being misjudged when $grid~size=0.5m$.
    We can also find from the bottom-right of the figure that when
    $grid~size=0.5m$, some object masks even become irregular, 
    which are supposed to be rectangles (\eg the orange and red ones).
    }
    \label{fig:gtcol}
\end{figure*}

\subsection{Main Results}
We conduct some ablation experiments 
in Table~\ref{performance}
to analyze the impact of 
velocity, acceleration, trajectories, and high-level command information to the performance of our model. 
We gradually add the acceleration, velocity, and high-level command information 
to the input, the average L2 error and 
collision rate continually decrease
from 0.97m to 0.29m, and 
0.49\% to 0.19\%.
It is worth mentioning that the collision rate of our method is not as low as some perception-based methods. We believe this is due to the insufficient decision-making process resulting from the mere fitting of motion information, which raises safety concerns.
Besides, the difference of average collision rates between our method and the others (\eg{0.19\% v.s. 0.14\%}) represents only about $2\sim3$ samples in all 4819 ones. 
We believe that more discriminative testing scenarios are needed to better demonstrate the advantages of perception-based methods.

\section{Analysis and Discussion}

\subsection{Trajectory Distribution of nuScenes}
This sub-section mainly analyzes the distribution of 
the ego vehicle's states on
the nuScenes training set from two perspectives: 
trajectory points in the future 3s, heading and curvature angles. 

\myPara{Trajectory Points} 
We plot all future 3s trajectory points in 
the training set in Figure~\ref{fig:Distribution} (a). 
It can be seen from the figure that the trajectories
are largely concentrated in the middle part (go straight), 
and the trajectories are mainly straight lines, or curves with very small curvatures.

\myPara{Heading and Curvature Angles}
The heading angle indicates the future driving direction relative to the current time, 
while the curvature angle reflects the vehicle's turning rate. 
As illustrated in Figure~\ref{fig:Distribution} (b) and (c), nearly 70\% of the heading angles and curvature angles lie within the ranges of $-0.2$ to $0.2$ and $-0.02$ to $0.02$ radians, respectively.
This finding is consistent with the conclusion drawn from the distribution of trajectory points.

Based on the above analysis on the distributions of the trajectory points,
heading and curvature angles, we argue that in the nuScenes training set, 
ego vehicles tend to move forward along straight lines and 
at small angles during driving in short-time horizons.

\subsection{Collision in Ground Truth}
When calculating the collision rate, 
the common practice of existing methods is to
first project objects such as vehicles and pedestrians into the bird's-eye-view (BEV) space 
and then convert them into occupied regions in the occupancy map,
where loss of precision occurs.
After this process, we find that a small fraction of ground-truth 
trajectory samples (about 2\%) also overlap with obstacles in the occupancy grid, 
resulting in collisions being falsely detected. 

However, the ego vehicle does not actually collide with any other objects while collecting data annotations. 
This anomaly is due to employing an occupancy map with a relatively large grid size, 
leading to false collisions when the ego vehicle approaches certain objects, 
\eg smaller than the size of a single occupancy map pixel.
We show an example of this phenomenon in Figure~\ref{fig:gtcol}, 
together with the collision detection results of ground-truth trajectories 
for two different grid sizes. 
Under a smaller grid size (0.1m) shown in the middle-right,
the evaluation system correctly identifies the ground-truth trajectory as not colliding, 
but under a larger grid size (0.5m) in the bottom-right, a false collision detection occurs.

After observing the impact of occupancy grid size on trajectory collision detection, 
we test with a grid size of 0.6m. 
The nuScenes training set has 4.8\% collision samples, while the validation set has 3.0\%. 
It is worth mentioning that when we previously use a grid size of 0.5m, 
only 2.0\% of the samples in the validation set are misclassified as collisions. 
This proves once again that the current method for determining the collision rate 
is not robust and accurate enough.

\section{Conclusion and Limitations} 
In this paper, we rethink the conventional evaluation metrics used in end-to-end autonomous driving by employing a simple MLP-based model that solely relies on physical states without any visual or point cloud perception as input. 
Despite its simplicity and the absence of perceptual information, our model achieves similar performance on the nuScenes dataset, suggesting that the current evaluation metrics may not adequately capture the superiority of different methods.

\vspace{.1cm}
\myPara{Limitations}
The primary objective of this paper is to present our observations rather than propose a new model. 
Our findings demonstrate the potential limitations of the current evaluation scheme on the nuScenes dataset.
Although our model performs well within the confines of the nuScenes dataset, we acknowledge that it is merely an impractical toy incapable of functioning in real-world scenarios. 
Driving without any knowledge beyond the ego vehicle's states is an insurmountable challenge. 

We believe that perception-based planning methods will ultimately become the solution to autonomous driving, producing much safer trajectories than our toy model.
We hope that our insights will stimulate further research in the field, encouraging a reevaluation and enhancement of the planning task for end-to-end autonomous driving.

{\small
\bibliographystyle{ieee_fullname}
\bibliography{egbib}
}

\end{document}